\documentclass[letterpaper]{article} 
\usepackage[preprint]{aaai2027} 
\usepackage[hyphens]{url}  
\usepackage{graphicx} 
\usepackage{natbib}  
\usepackage{caption} 
\usepackage{algorithm}
\usepackage{algorithmic}

\usepackage{newfloat}
\usepackage{listings}

\usepackage{xcolor}    
\usepackage{enumitem}  
\usepackage{amsmath}
\usepackage{tcolorbox}
\usepackage{booktabs}
\usepackage{pifont}
\usepackage[table]{xcolor}
\definecolor{tablegray}{gray}{0.92}
\newcommand{\cmark}{\ding{51}}
\newcommand{\xmark}{\ding{55}}
\newcommand{\pmark}{$\triangle$}
\usepackage{multirow} 

\tcbuselibrary{skins}      
\tcbuselibrary{breakable}  

\definecolor{promptheader}{RGB}{52, 73, 94}   
\definecolor{promptbody}{RGB}{248, 249, 250}   
\DeclareCaptionStyle{ruled}{labelfont=normalfont,labelsep=colon,strut=off} 
\floatstyle{ruled}
\newfloat{listing}{tb}{lst}{}
\floatname{listing}{Listing}

\usepackage{booktabs}

\title{TMCS: Tool-Grounded Multi-Agent Reasoning for Compositional Chemical Problem Solving}
\author{
Shengqin Wang\equalcontrib,
Jie Jin\equalcontrib,
Yu Cheng,
Yihang Chen,
Weilin Luo,
Yuan Xie,
Zhizhong Zhang
}

\affiliations{
East China Normal University,
Shanghai Innovation Institute,
University College London,
Huawei Noah's Ark Lab
}

\begin{document}

\maketitle

\begin{abstract}
Despite the promise of Large Language Models (LLMs) in computational chemistry, rigorous combinatorial chemistry problems remain difficult because they require quantitatively constrained molecular modification, candidate validation, and systematic revision after failed attempts. Existing tool-augmented chemical agents demonstrate useful planning and tool use, but they rarely provide a unified loop for property-driven molecular optimization and workflow-level composition. To bridge this gap, we propose Tool-Grounded Multi-Agent Reasoning for Compositional Chemical Problem Solving (TMCS), a step-by-step multi-agent framework that formalizes chemical problem solving as an interpretable, tool-augmented workflow. At the task level, specialized agents leverage external tools, few-shot trajectory memory, and structured reflection to iteratively refine solutions. At the workflow level, TMCS chains generation, understanding, editing, description, and optimization into a closed-loop pipeline. Evaluations across multiple chemical tasks demonstrate that TMCS consistently enhances chemical reasoning across both open- and closed-source base models, achieving state-of-the-art performance.
\end{abstract}

\section{Introduction}

Large Language Models (LLMs) and specialized Chemical Language Models (CLMs) have recently emerged as a transformative force in accelerating drug discovery and materials science~\cite{bai2025intern, ma2024llm, shojaee2025llm, hatakeyama2023prompt, xia2025nature, han2025generalist, zhang2025scientific, xia2022systematic, lv2024navigating, zhang2024chemllm, tan2025chemmllm, zhao2024chemdfm1, jiang2025chem3dllm, ross2022large, chithrananda2020chemberta, ahmad2022chemberta}. However, rigorous chemical problem-solving—such as drug design—is inherently procedural rather than instantaneous. Such tasks are rarely resolved through a one-shot prediction; instead, they necessitate a logical sequence of operations, including structural understanding, targeted modification, and iterative refinement~\cite{ock2026large}. This stepwise complexity underscores a critical yet overlooked requirement: true chemical intelligence must be assessed not solely by the final output, but by the system's capacity for transparent, intermediate logical reasoning~\cite{hao2026beyond}.
\begin{figure}[t]
  \includegraphics[width=\columnwidth]{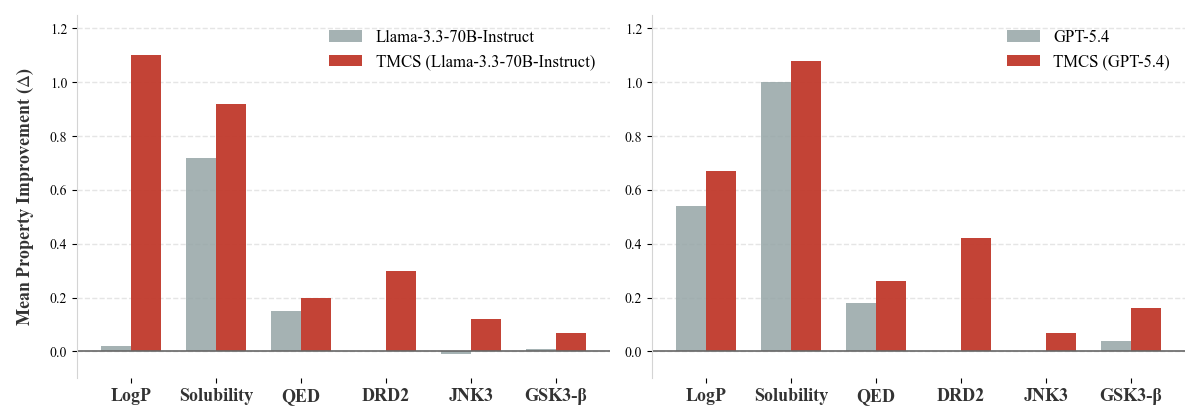}
  \caption{\textbf{Mean property improvement ($\Delta$) across six molecular optimization objectives.} The grouped bar charts compare the proposed TMCS framework against its corresponding generative baselines: Llama-3.3-70B-Instruct (left) and GPT-5.4 (right). TMCS consistently achieves superior $\Delta$ values across both physicochemical constraints and bioactivity targets. Higher positive values indicate successful property enhancements, whereas values near or below zero signify property degradations.}
  \label{fig:1}
\end{figure}
Existing paradigms face bottlenecks that hinder this procedural reasoning~\cite{wang2025chem,bai2025intern,li2026agentic,solovev2025madd,lee2026rag,liu2024drugagent}. First, at the individual task level, most methods treat chemical problems as direct, end-to-end predictions. Large language models (LLMs) and diffusion-based approaches map an input directly to an output relying on parameterized, black-box knowledge. This lacks reasoning transparency: structural changes are not visible step-by-step, making it difficult to interpret why a candidate succeeds or fails under fine-grained chemical constraints. Second, at the execution level, standard models exhibit what we term execution rigidity. Even when basic external tools are introduced, they rarely revise their strategy after a failed attempt, lack explicit reflection, and cannot reuse prior successful trajectories. Finally, at the workflow level, existing models often operate in knowledge isolation, executing tasks like generation, understanding, and optimization as disjointed processes rather than a continuous, cohesive pipeline.

The core difficulty is particularly acute for molecular optimization. A useful drug-design assistant must not only generate a syntactically valid molecule, but also determine whether a local edit actually improves logP, QED, solubility, or target-specific bioactivity while preserving a meaningful scaffold. Pure LLM prompting can express chemical intuition, yet it cannot reliably observe the quantitative consequence of each edit. Conversely, standalone tools can score or validate candidates but do not decide how to revise a failed design. This gap motivates a framework in which the LLM proposes chemically plausible actions, external tools provide objective feedback, and a bounded reflection loop revises the strategy only when the current trajectory is invalid or insufficiently improved.

To address these limitations, we propose TMCS, an interpretable, stepwise multi-agent framework for chemical problem solving. Built on the principle that chemical reasoning requires structured execution rather than simple black-box scaling, TMCS shifts the paradigm from one-shot generation to a closed-loop, tool-grounded refinement process. To achieve this, we introduce a dual-level optimization strategy.
First, we introduce the Task-Level Iterative Refinement: rather than stopping after a single prediction, TMCS decomposes individual tasks into executable stages. Specialized agents handle understanding, editing, and description, while external tools provide deterministic property validation. A reflection module monitors intermediate outputs and triggers strategy revision upon failure, guided by a few-shot trajectory memory that preserves reusable patterns from successful cases. As shown in Figure~\ref{fig:1}, our framework achieves significant improvements in property optimization for both open-source and closed-source models.

Second, we introduce the Workflow-Level Composition: moving beyond isolated execution, TMCS chains these diverse, optimized tasks into a unified pipeline. By passing context sequentially--from molecular generation, to structural understanding, and finally to targeted optimization--the framework ensures that insights gained in early stages directly inform downstream reasoning. This dual-level design improves interpretability, as every edit is linked to a specific structural rationale and validation result. 

Extensive evaluations show that TMCS substantially improves both open-source and closed-source base models, achieving state-of-the-art performance. Component, initialization, robustness, and cost analyses further indicate that the gains arise from verifiable tool-grounded feedback and bounded reflection rather than from unconstrained extra computation alone.

Our main contributions are summarized as follows:
\begin{itemize}
\item We propose \textbf{TMCS}, a stepwise multi-agent reasoning framework that transitions chemical problem solving from opaque, one-shot predictions to a transparent, closed-loop refinement process.
\item We design a \textbf{dual-level execution architecture}: at the task level, it enhances structural understanding and optimization through tool-use, few-shot reasoning trajectories, and reflection; at the workflow level, it composes these tasks into a unified, sequential pipeline.
\item We conduct extensive experiments showing that TMCS broadly enhances both open and closed-source LLMs. The framework yields substantial performance enhancements, achieving state-of-the-art results in chemical reasoning tasks and highlighting the superiority of structured execution.
\end{itemize}
\begin{figure*}[t]
  \includegraphics[width=\linewidth]{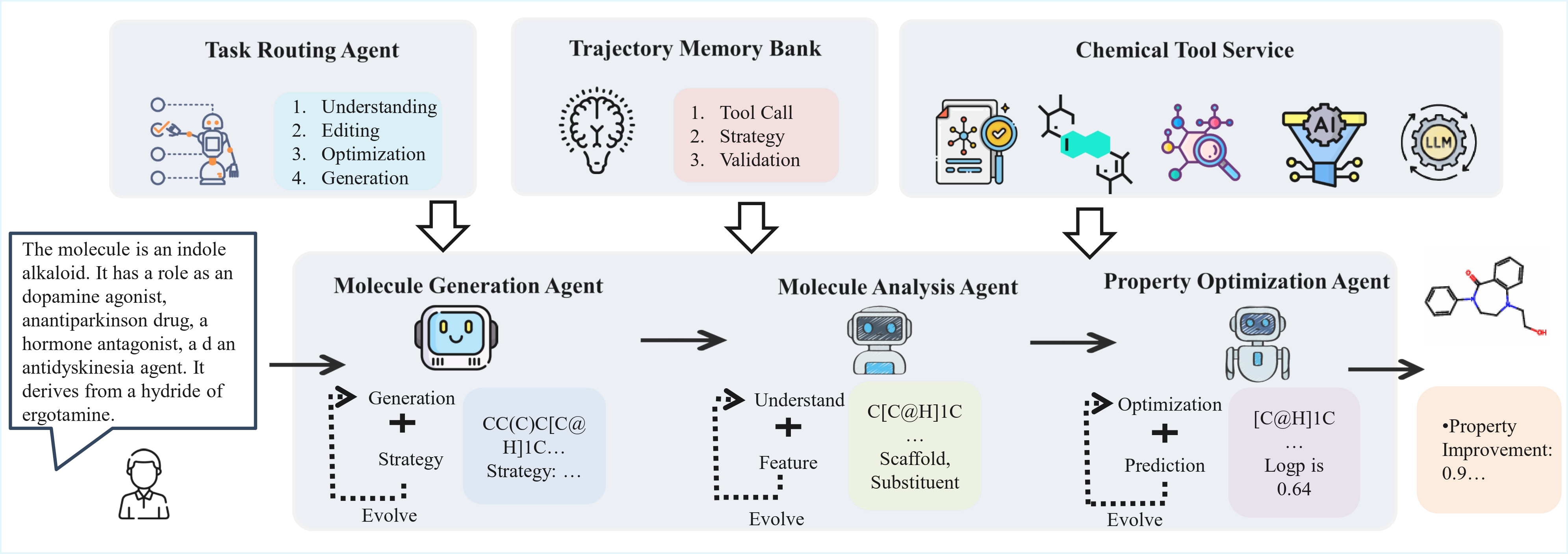} 
  \caption {The overall macroscopic architecture of the TMCS framework. It illustrates the compositional chemical pipeline supported by the Task Router, Memory Bank, and Tool Service.}
  \label{fig:22}
\end{figure*}
\begin{figure}[t]
  \includegraphics[width=\columnwidth]{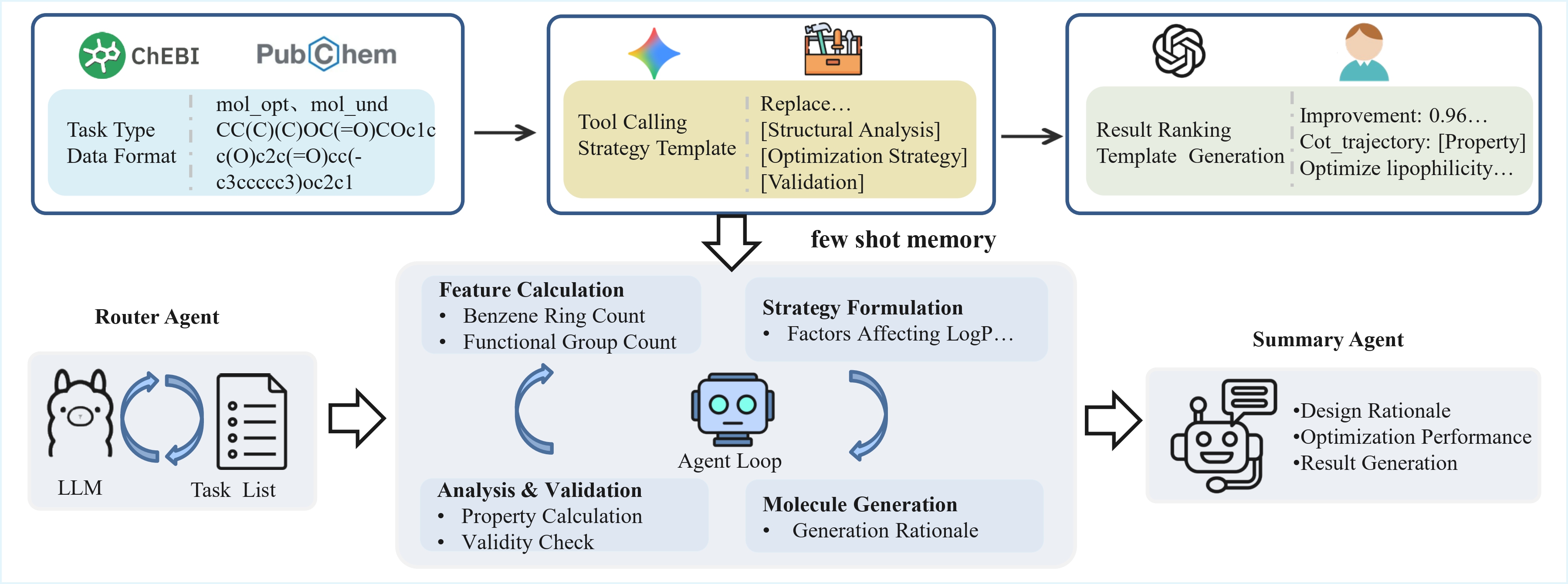}
  \caption{The task-level tool-grounded Agent Loop and memory construction process. It details how raw chemical data is transformed into few-shot memory to guide the self-evolving refinement cycle.}
  \label{fig:33}
\end{figure}

\section{Related work}
\subsection{Chemical Reasoning Models and Evaluation}
Prior work shows that many traditional evaluations emphasize factual recall and short-answer prediction, making it hard to distinguish knowledge retrieval from genuine chemical reasoning ~\cite{hao2026beyond,guo2025deepseek} . To address this gap, ChemCoTBench formulates molecular add/delete/substitute operations as reasoning primitives, promoting a modular evaluation paradigm that better matches chemist workflows~\cite{hao2026beyond}. Meanwhile, chemistry foundation models such as ChemLLM, ChemDFM, and LlasMol have improved performance on molecular and reaction tasks~\cite{zhang2024chemllm,zhao2024chemdfm,yu2024llasmol}, but they still face limitations in cross-task robustness, reasoning consistency, and interpretability\cite{wang2025chem}. More recent methods, such as Chem-R, introduce protocol-guided distillation, process-level supervision, and reinforcement-style optimization to improve the reliability of long-horizon chemical reasoning~\cite{wang2025chem,bai2025intern}

\subsection{Multi-Agent Systems for Drug Discovery}
Another major direction is tool-augmented multi-agent systems for drug discovery. Frameworks such as ChemCrow, ChemAgent, CACTUS, and DrugAgent combine LLM planning with chemistry tools, establishing increasingly mature workflows for molecular understanding, reaction analysis, and candidate generation~\cite{bran2023chemcrow,tang2025chemagent,liu2024drugagent}. MADD further extends this paradigm toward end-to-end hit identification through task decomposition and agent orchestration~\cite{solovev2025madd}. However, closed-loop molecular optimization remains underexplored. It requires quantitatively constrained structural edits, multi-property balancing, candidate validation, and systematic revision after failed modifications. Existing systems provide limited support and suffer from scarce optimization trajectories, tool-chain error accumulation, and cost and privacy constraints~\cite{solovev2025madd,bran2023chemcrow}. ChemCRAFT partially addresses these issues by delegating deterministic computation to external tools while retaining high-level decisions within the model~\cite{li2026agentic}. This gap motivates TMCS, which integrates structured reasoning, tool-grounded validation, and multi-agent execution for reliable molecular optimization.

Compared with these systems, TMCS focuses specifically on property-driven molecular optimization and its composition with upstream generation and understanding tasks. As summarized in Table~\ref{tab:positioning}, TMCS integrates tool use, memory, and reflection into an optimization-oriented protocol in which candidate molecules are generated, externally validated, scored by property oracles, and revised through bounded feedback.

\begin{table}[t]
\centering

\setlength{\tabcolsep}{3.5pt}
\renewcommand{\arraystretch}{1.2}

\begin{tabular}{@{}lcccc@{}}
\toprule
\textbf{Method}
& \shortstack{\textbf{Bounded}\\\textbf{reflection}}
& \shortstack{\textbf{Task}\\\textbf{routing}}
& \shortstack{\textbf{Closed-loop}\\\textbf{optimization}}
& \shortstack{\textbf{Multi}\\\textbf{task}} \\
\midrule

ChemCrow
& \xmark & \pmark & \xmark & \xmark \\

DrugAgent
& \xmark & \pmark & \pmark & \xmark \\

MADD
& \pmark & \cmark & \pmark & \xmark \\

ChemCRAFT
& \pmark & \xmark & \pmark & \xmark \\

\textbf{TMCS}
& \textbf{\cmark}
& \textbf{\cmark}
& \textbf{\cmark}
& \textbf{\cmark} \\

\bottomrule

\end{tabular}
\caption{Capability comparison with representative tool-augmented chemical
agents. \cmark: full support; \pmark: partial support; \xmark: absent or not
a primary focus.}
\label{tab:positioning}
\end{table}

\section{Method}

\subsection{Overview of the TMCS Framework}

We formulate chemical problem solving as a structured dual-level reasoning process that maps an optional combined query $x = (x_{\text{text}}, x_{\text{mol}})$ to an optimized output $y$. Rather than relying on a single black-box prediction, our framework decomposes complex tasks through a holistic architecture. As illustrated in Figure~\ref{fig:22}, the TMCS framework is supported by three foundational pillars: the Task Routing Agent, the Trajectory Memory Bank, and the Chemical Tool Service.

Given an input $x$, the framework operates across two granularities. At the \textbf{Workflow-Level}, macro-tasks are sequentially chained (e.g., generation $\rightarrow$ analysis $\rightarrow$ optimization). At the \textbf{Task-Level}, each specialized agent assigned by the Task Router executes a self-evolving loop. Formally, the end-to-end execution can be denoted as the composition of $K$ specialized agents:
\begin{equation}
y_{\text{final}} = \left( \prod_{k=1}^{K} A_k \right) (x; \mathcal{M}, \mathcal{T})
\end{equation}
where $\mathcal{M}$ represents the dynamic trajectory memory and $\mathcal{T}$ represents the integrated tool suite. This compositional design guarantees both macroscopic task coherence across diverse reasoning stages and microscopic structural precision at the atomic level.

\subsection{Chemical Tool Service}

To ground LLM reasoning in executable chemistry, TMCS instantiates a hybrid tool stack $\mathcal{T}$ that normalizes all tool invocations into JSON-compatible program calls~\cite{li2026agentic,wang2025chem}. The stack supports structural validation, deterministic property calculation, and constrained molecular editing. Candidate molecules must be chemically parseable, while editing operations are additionally verified through SMARTS-count changes to ensure that the requested functional group is added, removed, or replaced exactly as specified. Physicochemical descriptors, molecular similarities, structural patterns, and pharmacological properties are computed using deterministic backends or aligned predictive models rather than inferred directly by the LLM.

For molecular design and captioning, TMCS employs a specialized model derived from Chem-R as a replaceable generation tool. Generated structures and guarded LLM fallback outputs are subjected to the same validity and edit-consistency checks before being accepted into the workflow. This design combines deterministic chemical execution with flexible generation while maintaining a strict validation boundary. A complete description of the tools, interfaces, and functional roles is provided in supplementary materials.

\subsection{Trajectory Memory Construction and Retrieval}
\label{sec:traj_memory}

To mitigate zero-shot instability during complex chemical reasoning, we introduce an offline-constructed Few-Shot Trajectory Memory Bank, denoted by $\mathcal{M}$, which supplies executable priors during inference. As shown in supplementary materials, we provide additional specifications regarding this section.

\paragraph{Trajectory Construction.}
As shown in Figure~\ref{fig:33}, we construct our instruction dataset following~\cite{li2026agentic} through a three-step process. First, we collect high-fidelity molecular pairs, property annotations, and functional descriptions from databases like ChEBI and PubChem, standardizing them into a uniform SMILES-centric format based on task types. Second, we integrate these data instances with strategy templates that explicitly decompose chemical transformations into logical, tool-executable cognitive steps (i.e., structural analysis, optimization strategy, and validation). Finally, we employ large language models and external tools to infer initial reasoning trajectories.

\paragraph{Structured Memory and Deterministic Retrieval.}
We construct a task-bounded memory bank that encapsulates structured, deterministically executable exemplars (e.g., tool-call records) across molecular understanding, editing, and optimization tasks. To ensure rigorous structural alignment, we employ a deterministic, symbolic indexing mechanism based on exact key matching directed by task types. By directly matching exemplars to the specific task category, we guarantee that the retrieved trajectories are strictly compatible with the active query and immediately applicable to the current chemical problem. Each specific task is associated with its corresponding few-shot demonstrations. Furthermore, to rigorously prevent train-test contamination, the memory bank is constructed entirely offline, remains frozen during evaluation, and enforces strict structural disjointness from the test sets through canonical de-duplication and topological distance thresholding.

\subsection{Task-Level: Tool-Grounded Agent Loop}

Upon receiving a specific task assignment from the Router Agent, the system initiates the core \textit{Agent Loop}. As shown in Figure~\ref{fig:33}, taking the \textbf{property optimization task} as a representative paradigm, the internal loop dynamically orchestrates a  sequence of actions  based on the current context.

\paragraph{Chemical Agent Loop.}
The chemical agent loop proceeds through four integrated phases: (1) \textit{Feature Calculation}, where the agent invokes $\mathcal{T}_{\text{calc}}$ to extract physicochemical features (e.g., functional group counts and molecular weight) from the current molecule $y^{(t-1)}$; (2) \textit{Strategy Formulation}, where guided by the extracted features, target constraints, and retrieved memory $\mathcal{M}$, the agent articulates a targeted structural modification strategy $z^{(t)}$; (3) \textit{Generation \& Editing}, where the agent utilizes $\mathcal{T}_{\text{edit}}$ or $\mathcal{T}_{\text{gen}}$ to actualize the proposed strategy, yielding a candidate molecule $y^{(t)}$ along with its generation rationale; and (4) \textit{Analysis \& Validation}, where the candidate is rigorously evaluated via $\mathcal{T}_{\text{valid}}$ and $\mathcal{T}_{\text{calc}}$ to return an objective feedback vector $\mathbf{r}^{(t)}$.

\paragraph{Refinement and Summarization.}
Upon receiving the feedback vector, the agent performs a reflection operation to analyze the current solution's efficacy. In our implementation, molecular optimization uses at most three refinement rounds and generates three candidates per round. The best valid candidate is selected according to property improvement and scaffold similarity; if no candidate improves the objective, the next prompt receives explicit feedback describing the best failed attempt and whether more conservative or more aggressive edits are needed. The reflection module classifies errors into format, structural, counting, logic, hallucination, or unknown categories, then appends a targeted correction prompt rather than an open-ended instruction to ``think again.'' This self-evaluation drives further iterative refinement, ultimately culminating in the final task solution and a comprehensive analytical summary report.

\subsection{Workflow-Level: Compositional Chemical Pipeline}

For multifaceted, real-world chemical challenges (e.g., designing a targeted drug from a raw textual description), a single autonomous loop is insufficient due to context window dilution and error accumulation. At the workflow level, TMCS chains diverse, specialized agents into a macroscopic pipeline.

As shown in Figure~\ref{fig:22} , the workflow initiates with a \textbf{Molecule Generation Agent} translating textual descriptions (e.g., ``an indole alkaloid derivative'') into initial structural scaffolds. The finalized summary of this agent is seamlessly handed over as the initial context to a \textbf{Molecule Analysis Agent}, which isolates modifiable substituents. Finally, the \textbf{Property Optimization Agent} receives this focused analysis to perform the targeted refinement loop discussed above. This compositional architecture ensures that the cognitive load is compartmentalized, preventing the cascading of hallucinations while preserving high-fidelity reasoning across complex chemical workflows.

\begin{figure}[t]
  \includegraphics[width=\columnwidth]{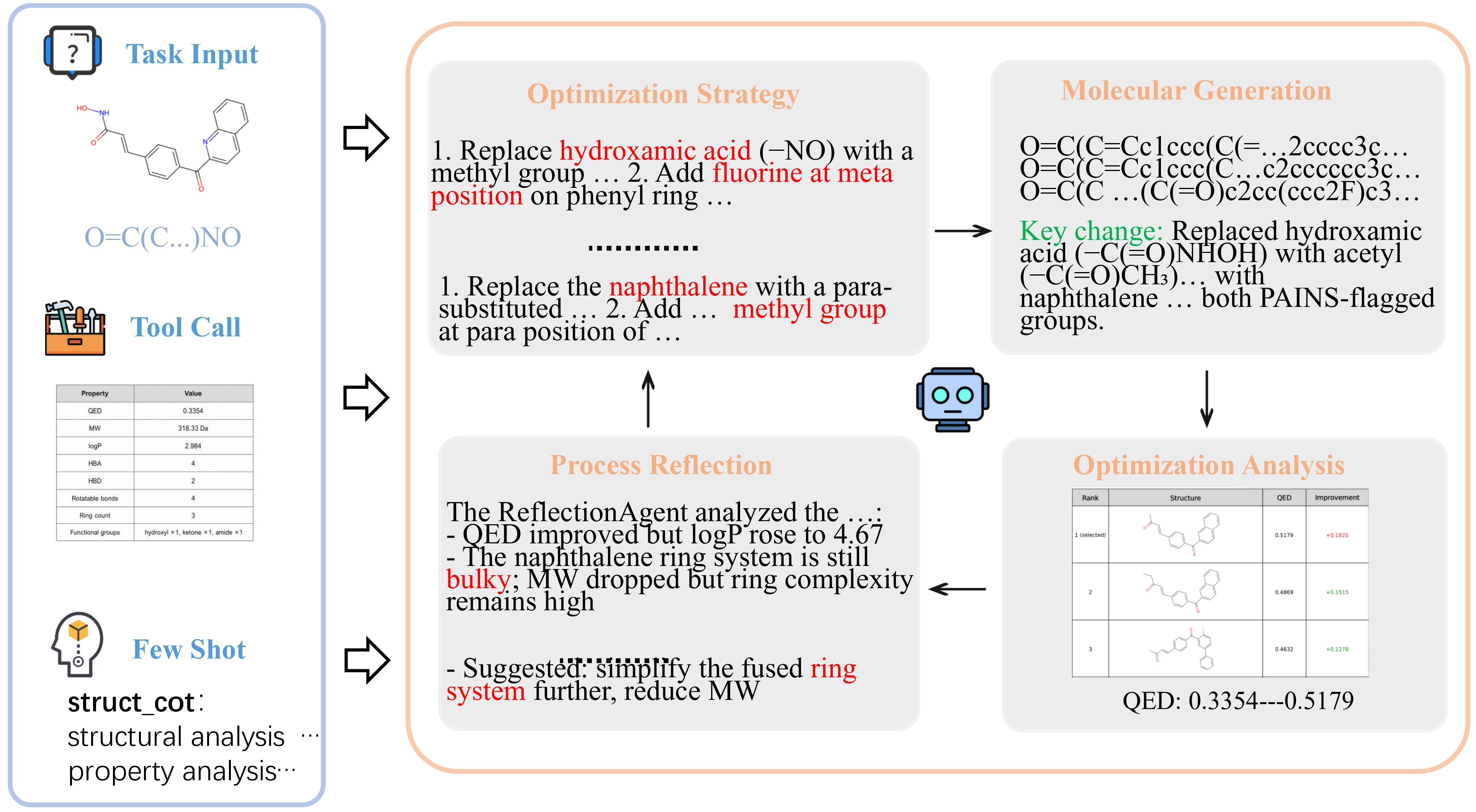}
  \caption{Example of QED attribute optimization.}
  \label{fig:caseqed}
\end{figure}

\section{Experiments}
\subsection{Experimental Setup}

\textbf{Benchmark.} To comprehensively validate our framework, we conducted evaluations across a broad spectrum of tasks. At the task level, we evaluated drug molecule design and molecular description from ChemLLMBench~\cite{guo2023can}, alongside molecular understanding, molecular editing, and property optimization tasks from ChemCoTBench~\cite{hao2026beyond}. At the workflow level, we utilized the molecular design task~\cite{guo2023can} as an initialization point to sequentially drive downstream optimization tasks and evaluate the optimization performance across six distinct molecular properties.

\textbf{Evaluation Metrics.} For rigorous and fair benchmarking, we follow standard protocols from prior work~\cite{guo2023can, hao2026beyond}. Molecular design is evaluated by exact SMILES match, while molecule captioning uses BLEU-4. Molecular understanding evaluates functional-group and ring counting with Mean Absolute Error (MAE), Murcko scaffold extraction with Tanimoto similarity, complex-ring detection with accuracy, and SMILES equivalence as a binary classification task using accuracy. Molecular editing is measured by Pass@1, indicating whether the edited structure strictly follows the instruction, whereas molecular optimization is assessed by average property improvement. Direct LLM baselines receive task-specific optimization prompts and the same property oracles as TMCS; complementary ablations and cost analyses further distinguish architectural gains from those due to additional inference budget. The LLM parameters are set to `temperature = 0.3` and max tokens are 512, with up to eight retries upon failure.


\textbf{Baselines.} To establish a rigorous and comprehensive comparison, we categorize our baselines into three distinct paradigms. For direct reasoning, we adopt the direct inference baselines established in ChemCoTBench to evaluate standard prompting capabilities~\cite{grattafiori2024llama,hurst2024gpt,guo2025deepseek,comanici2025gemini,singh2025openai}. For domain-specific systems, we select Chem-R, ether0~\cite{narayanan2026training}, and ChemCRAFT as our primary comparators. Finally, to validate the generalizability of our proposed framework at both the task and workflow levels, we deploy TMCS on top of leading foundational models, utilizing both open-source (specifically \texttt{Llama-3.3-70B-Instruct}) and closed-source (specifically \texttt{GPT-5.4}) LLMs as our base models.

\subsection{Main Results}

\definecolor{tablegray}{gray}{0.85}

\begin{table*}[t]
\centering
\setlength{\tabcolsep}{4pt}

\begin{tabular}{lcccccccccc}
\toprule
Methods & FG$\downarrow$ & Ring$\downarrow$ & Murcko$\uparrow$ & Ring-sys$\uparrow$ & Eq$\uparrow$ & Add & Delete & Sub & Design & Capt \\
\midrule

\rowcolor{tablegray}
\multicolumn{11}{c}{\textit{General Models}} \\

GPT-4o                 
& 0.17 & 1.35 & 0.21 & 80.0 & 72 & 80 & 80 & 65.0 & 0.07 & 0.01 \\

DeepSeek-R1            
& 0.27 & 1.55 & 0.34 & 45.0 & 65 & 70 & 70 & 68.3 & 0.22 & 0.04 \\

GPT 5.4                
& 0.29 & 1.05 & 0.42 & 45.0 & 84 & 80 & 80 & 88.3 & 0.19 & 0.13 \\

Gemini-2.5-pro         
& 0.11 & 0.60 & 0.51 & 87.5 & 82 & \textbf{100} & 85 & 81.7 & 0.29 & 0.04 \\

Llama-3.3-70B-Instruct 
& 0.52 & 1.80 & 0.12 & 68.3 & 67 & 60 & 80 & 50.0 & 0.03 & 0.02 \\

\midrule

\rowcolor{tablegray}
\multicolumn{11}{c}{\textit{Chemical Models}} \\

ChemCRAFT              
& 0.03 & 0.15 & 0.57 & \textbf{100.0} & 97 & 85 & 95 & 80.0 & - & - \\

ether0                 
& - & 0.35 & - & - & 63 & 94 & 76 & 78.0 & 0.30 & 0.03 \\

Chem-R                 
& - & - & - & - & - & - & - & - & 0.42 & 0.41 \\

BioMedGPT              
& 1.60 & 2.43 & 0.18 & 53.3 & 39 & 10 & 12 & 10.0 & - & - \\

BioMistral             
& 1.00 & 1.85 & 0.04 & 32.5 & 50 & 0 & 10 & 0.0 & - & - \\

TMCS (70B)             
& 0.31 
& \textbf{0.00} 
& \textbf{1.00} 
& \textbf{100.0} 
& \textbf{99} 
& 90 
& 90 
& 95.0 
& 0.44 
& 0.48 \\

TMCS (GPT 5.4)         
& \textbf{0.00} 
& \textbf{0.00} 
& \textbf{1.00} 
& \textbf{100.0} 
& \textbf{99} 
& 95
& \textbf{100} 
& \textbf{96.7} 
& \textbf{0.47} 
& \textbf{0.49} \\

\bottomrule
\end{tabular}

\caption{Results on individual chemical tasks. 70B represents Llama-3.3-70B-Instruct, $\downarrow$ indicates lower is better, while $\uparrow$ indicates higher is better.}
\label{tab:tasklevel}
\end{table*}

\definecolor{tablegray}{gray}{0.85}
\begin{table*}[t]
\centering

\setlength{\tabcolsep}{4.5pt} 

\begin{tabular}{l cccccccccccc}
\toprule
\multirow{2}{*}{\textbf{Methods}} 
& \multicolumn{2}{c}{\textbf{LogP}} 
& \multicolumn{2}{c}{\textbf{Solubility}} 
& \multicolumn{2}{c}{\textbf{QED}} 
& \multicolumn{2}{c}{\textbf{DRD2}} 
& \multicolumn{2}{c}{\textbf{JNK3}} 
& \multicolumn{2}{c}{\textbf{GSK3-$\beta$}} \\

\cmidrule(lr){2-3} 
\cmidrule(lr){4-5} 
\cmidrule(lr){6-7} 
\cmidrule(lr){8-9} 
\cmidrule(lr){10-11} 
\cmidrule(lr){12-13}

& $\Delta$ & SR 
& $\Delta$ & SR 
& $\Delta$ & SR 
& $\Delta$ & SR 
& $\Delta$ & SR 
& $\Delta$ & SR \\

\midrule

\rowcolor{tablegray}
\multicolumn{13}{c}{\textit{General Models}} \\

GPT-4o                 
& -0.09 & 37 
& 0.92 & 80 
& 0.13 & 70 
& 0.07 & 48 
& -0.02 & 30 
& -0.00 & 39 \\

DeepSeek-R1            
& 0.47 & 69 
& 0.80 & 80 
& 0.17 & 72 
& 0.12 & 62 
& -0.02 & 29 
& 0.01 & 41 \\

Gemini-2.5-pro         
& -0.22 & 76 
& 1.06 & 70 
& \textbf{0.28} & 84 
& 0.36 & 74 
& -0.02 & 35 
& 0.06 & 68 \\

Llama-3.3-70B-Instruct 
& 0.02 & 35 
& 0.72 & 81 
& 0.15 & 61 
& 0.00 & 31 
& -0.01 & 30 
& 0.01 & 40 \\

GPT-5.4                
& 0.54 & 83 
& 1.00 & 90 
& 0.18 & 91 
& 0.00 & 57 
& 0.00 & 39 
& 0.04 & 69 \\

\midrule

\rowcolor{tablegray}
\multicolumn{13}{c}{\textit{Chemical Models}} \\

ether0                 
& 0.00 & 0 
& 0.00 & 0 
& 0.00 & 0 
& 0.00 & 0 
& 0.00 & 0 
& 0.00 & 0 \\

Chem-R                 
& - & - 
& 0.34 & 83 
& - & - 
& 0.01 & 36 
& -0.02 & 24 
& -0.01 & 29 \\

BioMistral             
& 0.01 & 1 
& 0.24 & 6 
& 0.00 & 0 
& 0.00 & 1 
& -0.01 & 1 
& -0.01 & 0 \\

BioMedGPT              
& -0.36 & 17 
& 0.25 & 63 
& -0.29 & 7 
& -0.09 & 5 
& -0.11 & 6 
& -0.08 & 1 \\

TMCS (70B)             
& \textbf{1.10} & 96 
& 0.92 & 73 
& 0.20 & 80 
& 0.30 & 90 
& \textbf{0.12} & \textbf{72} 
& 0.07 & 49 \\

TMCS (GPT 5.4)         
& 0.90 & \textbf{98} 
& \textbf{1.09} & \textbf{99} 
& 0.25 & \textbf{98} 
& \textbf{0.42} & \textbf{92} 
& 0.07 & 61 
& \textbf{0.16} & \textbf{79} \\

\bottomrule
\end{tabular}

\caption{Optimization results on multiple molecular objectives. 
$\Delta$ is the mean property improvement, where a negative $\Delta$ indicates that most
optimizations are property degradations. SR denotes success rate (\%).}
\label{tab:tasklevel2}

\end{table*}

\textbf{Task-Level Benchmark Results.} 
Table~\ref{tab:tasklevel} shows that TMCS achieves strong gains across molecular understanding, editing, generation, and description, outperforming strong LLM baselines particularly on tasks requiring multi-step reasoning, precise structural manipulation, or description generation. These results demonstrate that the same stepwise reasoning architecture generalizes beyond a single optimization objective. They also justify a dedicated molecular optimization module: because text-to-molecule generation may produce plausible scaffolds but cannot ensure property improvement under structural constraints, TMCS uses generation only for initialization and applies tool-grounded optimization to obtain quantitatively improved candidates.



Table~\ref{tab:tasklevel2} summarizes the task-level benchmark performance on multiple optimization objectives. TMCS consistently improves property scores across LogP, solubility, QED, DRD2, JNK3, and GSK3-$\beta$. Compared with both open and closed baselines, our method exhibits stronger optimization ability and more stable success rates.
\begin{table*}[t]
\centering
\setlength{\tabcolsep}{3.5pt}
\begin{tabular}{ll cc cc cc cc cc cc}
\toprule
\multirow{2}{*}{\textbf{Model}} & \multirow{2}{*}{\textbf{Strategy}} & \multicolumn{2}{c}{\textbf{LogP}} & \multicolumn{2}{c}{\textbf{Solubility}} & \multicolumn{2}{c}{\textbf{QED}} & \multicolumn{2}{c}{\textbf{DRD2}} & \multicolumn{2}{c}{\textbf{JNK3}} & \multicolumn{2}{c}{\textbf{GSK3-$\beta$}} \\
\cmidrule(lr){3-4} \cmidrule(lr){5-6} \cmidrule(lr){7-8} \cmidrule(lr){9-10} \cmidrule(lr){11-12} \cmidrule(lr){13-14}
 & & $\Delta$ & SR & $\Delta$ & SR & $\Delta$ & SR & $\Delta$ & SR & $\Delta$ & SR & $\Delta$ & SR \\
\midrule
\multirow{2}{*}{70B} & Baseline & 0.152 & 41.0 & 0.004 & 15.0 & -0.003 & 16.0 & 0.000 & 17.0 & 0.001 & 15.0 & 0.002 & 14.0 \\
 & \textbf{TMCS (Ours)} & \textbf{1.056} & \textbf{89.0} & \textbf{0.416} & \textbf{65.0} & \textbf{0.123} & \textbf{81.0} & \textbf{0.094} & \textbf{69.0} & \textbf{0.028} & \textbf{68.0} & \textbf{0.085} & \textbf{74.0} \\
\midrule
\multirow{2}{*}{GPT-5.4} & Baseline & -0.417 & 82.0 & \textbf{1.305} & 80.0 & 0.095 & 78.0 & 0.002 & 42.0 & 0.004 & 30.0 & 0.003 & 35.0 \\
 & \textbf{TMCS (Ours)} & \textbf{1.101} & \textbf{98.0} & 1.062 & \textbf{92.0} & \textbf{0.178} & \textbf{94.0} & \textbf{0.148} & \textbf{80.0} & \textbf{0.054} & \textbf{96.0} & \textbf{0.124} & \textbf{90.0} \\
\bottomrule
\end{tabular}
\caption{Workflow-level optimization-stage results: comparison between the task-specific Pure LLM Baseline and TMCS across six molecular objectives. We report \textbf{$\Delta$} (mean net property change over all 100 cases, including positive and negative changes) and \textbf{SR} (success rate, \%). 70B represents Llama-3.3-70B-Instruct.}
\label{tab:optimization_main1}
\end{table*}

While the Llama-3.3-70B-Instruct model exhibits limited reasoning capabilities in direct inference tasks, the integration of the TMCS framework leads to substantial performance gains. Similarly, although powerful proprietary models like GPT-5.4 demonstrate baseline effectiveness in zero-shot settings, their performance is further amplified by TMCS. These results underscore that the TMCS framework provides a robust, universal mechanism for unlocking high-tier chemical reasoning, regardless of the underlying model's initial architecture.

\textbf{Workflow-Level Pipeline Performance.} To test whether the individual task gains can be composed into a stronger end-to-end system, we start from drug molecule generation and then perform downstream optimization. We compare the tool-enhanced pipeline against a single-turn, task-specific LLM baseline.

Table~\ref{tab:optimization_main1} presents the core results of this workflow-level optimization stage, whose initialization setting differs from the standalone task-level benchmark in Table~\ref{tab:tasklevel2}. We compare the \textbf{Tool-enhanced} TMCS framework against a \textbf{Pure LLM Baseline} using task-specific prompts but no external feedback during generation. The results demonstrate a significant performance gap, particularly in properties like DRD2 and JNK3, where the lack of quantitative feedback causes the pure LLM to struggle ($\sim$15\%--42\% success). In contrast, TMCS leverages tool-grounded feedback to achieve a substantial performance boost, reaching success rates as high as 98\%.

The  table reports two complementary workflow-level measures over all 100 molecules. \textbf{SR} is the fraction with a positive change, whereas $\Delta$ is the net mean change after summing both positive and negative changes. TMCS improves SR for all six objectives on both backbones, and its all-sample $\Delta$ is higher on all six objectives for 70B and on five of six objectives for GPT-5.4. This combination shows that TMCS is more stable across molecules rather than relying on a small number of large gains.

For \textbf{GPT-5.4}, the stability advantage is especially clear on the black-box objectives: TMCS raises SR from 42.0\% to 80.0\% on DRD2, from 30.0\% to 96.0\% on JNK3, and from 35.0\% to 90.0\% on GSK3-$\beta$; their net mean changes increase from 0.002 to 0.148, 0.004 to 0.054, and 0.003 to 0.124, respectively. Solubility is the only $\Delta$ exception: the baseline reaches 1.305 versus TMCS's 1.062, but TMCS still improves SR from 80.0\% to 92.0\%. Thus, the baseline's larger solubility magnitude does not overturn the broader evidence that tool-grounded feedback yields more reliable molecule-level improvements.


\begin{table}[t]
\centering

\begin{tabular}{@{}lcccc@{}}
\toprule
       & \multicolumn{2}{c}{LogP} & \multicolumn{2}{c}{DRD2} \\
\cmidrule(lr){2-3} \cmidrule(lr){4-5} 
Method & $\Delta$ & SR & $\Delta$ & SR \\
\midrule
\textbf{Ours} & \textbf{1.10} & \textbf{96} & \textbf{0.30} & \textbf{90} \\
w/o few-shot  & 0.96 & 88 & 0.27 & 84 \\
w/o tool      & 0.96 & 86 & 0.26 & 80 \\
\bottomrule
\end{tabular}
\caption{Ablation results on molecular optimization tasks. $\Delta$ is the mean property improvement, where a negative $\Delta$ indicates that most optimizations are property degradations, and SR denotes success rate.}

\end{table}

\begin{table}[t]
\centering

\begin{tabular}{@{}lc@{}}
\toprule
\textbf{ Setting} & \textbf{Solubility \textbf{$\Delta$}} \\
\midrule
Baseline                 & 0.72 \\
\midrule
\multicolumn{2}{@{}l}{\textit{Ablation on reflection rounds}} \\
w/o reflection           & 0.87 \\
1 round                  & 0.90 \\
5 rounds                 & 0.93 \\
\midrule
3 rounds (Ours)  & 0.92 \\
\bottomrule
\end{tabular}
\caption{Ablation results on reflection rounds for molecular optimization. \textbf{$\Delta$} denotes mean property improvement; TMCS uses three rounds by default.}
\label{tab:ablation1}
\end{table}

\begin{table}[htbp]
\centering

\footnotesize 
\setlength{\tabcolsep}{3pt} 
\begin{tabular}{lll ccc}
\toprule
\textbf{Model} & \textbf{Start Source} & \textbf{Strategy} & \textbf{DRD2} & \textbf{GSK3-$\beta$} & \textbf{JNK3} \\
\midrule
\multirow{4}{*}{GPT-5.4} & \multirow{2}{*}{Gen-Start} & Baseline & 40.9 & 31.9 & 32.9 \\  
 & & \textbf{TMCS} & \textbf{78.8} & \textbf{94.2} & \textbf{95.7} \\
\cmidrule{2-6}
 & \multirow{2}{*}{GT-Start} & Baseline & 44.1 & 41.9 & 23.3 \\
 & & \textbf{TMCS} & \textbf{82.4} & \textbf{80.6} & \textbf{96.7} \\
\bottomrule
\end{tabular}
\caption{Ablation results on initialization source for bioactivity tasks (DRD2, GSK3-$\beta$, JNK3).}
\label{tab:ablation4}
\end{table}
\subsection{Ablation Study}

\textbf{Component Ablation}
Table~\ref{tab:ablation1} shows the contribution of each component on two representative optimization objectives, LogP and DRD2. The full model achieves the best performance on both metrics. Removing memory or tools consistently degrades performance, confirming that the gains cannot be attributed only to extra LLM calls.

The results reveal three important trends. First, removing few-shot memory reduces both improvement and success rate, indicating that reusable trajectory patterns are helpful for selecting better edits. Second, removing tools causes a larger drop in success rate, suggesting that deterministic validation is important for producing reliable candidates.

Regarding the ablation of reflection frequency, as shown in Table~\ref{tab:ablation1}, removing reflection weakens the system's ability to escape local plateaus, which is especially visible in the optimization tasks that require iterative refinement. One round already improves over the direct baseline, while five rounds only marginally improve solubility $\Delta$ from 0.92 to 0.93 compared with the three-round TMCS setting. We therefore use three rounds as the default because it captures nearly all of the performance gain while avoiding the additional latency and token cost of longer reflection. For tasks such as molecular understanding, a single reflection step is sufficient to achieve optimal performance, so we restrict the reflection mechanism to a single iteration for these tasks.

Additionally, we provide a detailed statistical significance and robustness analysis in supplementary materials, followed by a computational cost and efficiency analysis of our framework in supplementary materials.

\textbf{Impact of Initialization Source.}
To further understand the robustness of TMCS, we investigate the impact of the initialization source on optimization success. As shown in Table~\ref{tab:ablation4}, we split the cases into those starting from generated molecules (\textit{Generated-Start}) and those falling back to ground-truth molecules (\textit{GT-Start}). 

A key finding is that pure LLM optimization is highly sensitive to the starting source: performance varies substantially between generated scaffolds and ground-truth scaffolds across DRD2, GSK3-$\beta$, and JNK3. TMCS remains consistently strong in both settings. When the initial molecule is produced by a general molecular generator, TMCS can still further optimize it; when the system starts from a ground-truth scaffold, TMCS also improves it. Thus, Chem-R-style initialization is useful but not the sole source of the final gain; the tool-grounded optimization loop supplies the decisive navigation signal.

\textbf{Efficiency Trade-off.}
The multi-agent framework intentionally spends more inference than single-turn prompting because it performs candidate generation, tool scoring, and reflection. However, this overhead is controlled rather than open-ended: molecular optimization is capped at three rounds with three candidates per round, deterministic tasks are routed directly to tools, and short tasks use at most one reflection. Supplementary materials quantifies this trade-off. Direct inference is cheaper, but it lacks the feedback channel needed for long-step optimization; TMCS pays additional tokens only for modules that expose intermediate chemical evidence. The reflection-round ablation further shows that extending the loop beyond the default budget produces negligible gains, supporting the chosen quality-efficiency balance.

\textbf{Case Study.}
To provide an implementation-level qualitative analysis, Figure~\ref{fig:caseqed} illustrates a three-round QED optimization trajectory driven by the proposed pipeline. As depicted, the system integrates the task prompt, deterministic tool feedback, and few-shot structural rationales to initialize the generation cycle. 

Starting from the initial compound ($s_0$, $\mathrm{QED}=0.3354$), Round 1 executes a localized structural edit---replacing the PAINS-flagged hydroxamic acid liability with an acetyl group. When the trajectory plateaus in Round 2, the ReflectionAgent actively intervenes. By analyzing the updated tool profiles, it explicitly diagnoses that although the QED improved, the naphthalene ring system remained overly bulky. 

Conditioned on these tool-grounded diagnostics, the agent dynamically refreshes its strategy for Round 3. It shifts from localized editing to a deeper scaffold simplification (e.g., replacing the fused bicyclic naphthalene motif with a simpler para-substituted variant). This case concretely demonstrates the efficacy of the multi-round reflective logic: it detects stagnation, performs precise credit assignment via explicit diagnostics, and successfully escapes local optima that single-pass generative models typically fail to overcome.
\section{Conclusion}
\label{sec:conclusion}

TMCS is a tool-grounded multi‑agent framework that reformulates chemical problem‑solving into an interpretable, iterative workflow, integrating generative modeling with deterministic chemical logic and achieving SOTA  that lets open‑weight models match or surpass closed‑source ones.

\bibliography{aaai2027}

\newpage
\appendix

\end{document}